\documentclass[letterpaper, 10 pt, conference]{ieeeconf}  

\IEEEoverridecommandlockouts                              

\usepackage{amsmath,amsfonts}
\usepackage{algorithmic}
\usepackage{algorithm}
\usepackage[caption=false,font=normalsize,labelfont=sf,textfont=sf]{subfig}
\usepackage{textcomp}
\usepackage{stfloats}
\usepackage{url}
\usepackage{graphicx}
\usepackage{cite}
\usepackage{cleveref}
\usepackage{tabularx}

\usepackage{amsmath,amssymb,amsfonts}
\usepackage{graphicx}
\usepackage{xcolor}
\usepackage{colortbl}
\usepackage{array}
\def\BibTeX{{\rm B\kern-.05em{\sc i\kern-.025em b}\kern-.08em T\kern-.1667em\lower.7ex\hbox{E}\kern-.125emX}}

\usepackage{times}
\usepackage{multirow}
\usepackage{booktabs}
\usepackage{amsfonts}
\usepackage{balance}
\title{\LARGE \bf Synesthesia of Vehicles: Tactile Data Synthesis from Visual Inputs}

\author{Rui Wang$^{1}$, Yaoguang Cao$^{2}$, Yuyi Chen$^{3}$, Jianyi Xu$^{4}$, Zhuoyang Li$^{5}$, Jiachen Shang$^{1}$, Shichun Yang$^{1}$%
\thanks{*Supported by National Key R\&D Program of China, No: 2022YFB3206600}%
\thanks{\textsuperscript{1}Rui Wang et al. are with Dept. of Transportation Science, Beihang Univ., Beijing, China, E-mail: \{bhwangr, by2213106, yangshichun\}@buaa.edu.cn}%
\thanks{\textsuperscript{2}Yaoguang Cao is with State Key Lab of Intelligent Transportation System, Beihang Univ., Beijing and Hangzhou International Innovation Institute, Beihang Univ., Hangzhou, China, E-mail: caoyaoguang@buaa.edu.cn }%
\thanks{\textsuperscript{3}Yuyi Chen are with Low-altitude Vehicle Systems Laboratory, Hangzhou International Innovation Institute, Beihang Univ., Hangzhou, China, E-mail: yychen@buaa.edu.cn }%
\thanks{\textsuperscript{4}Jianyi Xu is with Hangzhou International Innovation Institute, Beihang Univ., Hangzhou, China, E-mail: zy2457928.edu.cn }%
\thanks{\textsuperscript{4}Zhuoyang Li are with China Software Testing Center(Ministry of Industry and Information Technology  Software and Integrated Circuit Promotion Center), China, E-mail: liszzylzy@163.com }%
\thanks{\textsuperscript{1}Shichun Yang is the corresponding author.}}

\begin{document}

\maketitle
\thispagestyle{empty}
\pagestyle{empty}

\begin{abstract}
Autonomous vehicles (AVs) rely on multi-modal fusion for safety, but current visual and optical sensors fail to detect road-induced excitations which are critical for vehicles' dynamic control. Inspired by human synesthesia, we propose the Synesthesia of Vehicles (SoV), a novel framework to predict tactile excitations from visual inputs for autonomous vehicles. We develop a cross-modal spatiotemporal alignment method to address temporal and spatial disparities. Furthermore, a visual-tactile synesthetic (VTSyn) generative model using latent diffusion is proposed for unsupervised high-quality tactile data synthesis. A real-vehicle perception system collected a multi-modal dataset across diverse road and lighting conditions. Extensive experiments show that VTSyn outperforms existing models in temporal, frequency, and classification performance, enhancing AV safety through proactive tactile perception.
\end{abstract}

\section{INTRODUCTION}
Autonomous vehicles (AVs) are advancing rapidly through multi-modal fusion and AI algorithms, primarily relying on vision-based and optical sensors such as cameras and LiDARs, which excel in tasks like pedestrian detection and lane recognition \cite{AV1}. However, these non-contact sensors fail to capture vehicle dynamics induced by road surfaces, such as tire slip conditions or dynamic tire loads, and conventional state estimation methods often lack robustness under complex conditions \cite{frictionsurvey}.

Human skin contains mechanoreceptors that sense mechanical stimulis like pressure and deformation, enabling tactile perception of texture, roughness, and slip states \cite{humanskin}. Similarly, vehicle tires, the only components in direct contact with the road, reflect road-induced excitations through their stress-strain states. Recent developments in intelligent tires with embedded sensors have enhanced AVs with tactile perception, monitoring road conditions and tire states in real-time \cite{cyysurvey}. For example, Maurya et al. developed 3D-printed strain sensors for real-time load monitoring \cite{pvdfnc}, while Sun et al. used PVDF sensor arrays with machine learning to measure tire slip angles with high precision \cite{xuslip}.

Integrating tactile perception with visual data offers a more comprehensive environmental understanding for AVs. Shi et al. proposed a visual-tactile fusion framework for accurate road type identification \cite{shifusion}, and Wang et al. demonstrated its superiority over single-modality perception under varying illumination conditions \cite{wangfusion}. However, challenges persist: visual perception detects upcoming road conditions but not real-time excitations from road surfaces, while tactile perception captures current excitations without foresight. Eliminating the spatiotemporal disparities between visual and tactile modalities is a key issue in visual-tactile fusion research.

\begin{figure}[htp]
    \centering
    \includegraphics[width=8cm]{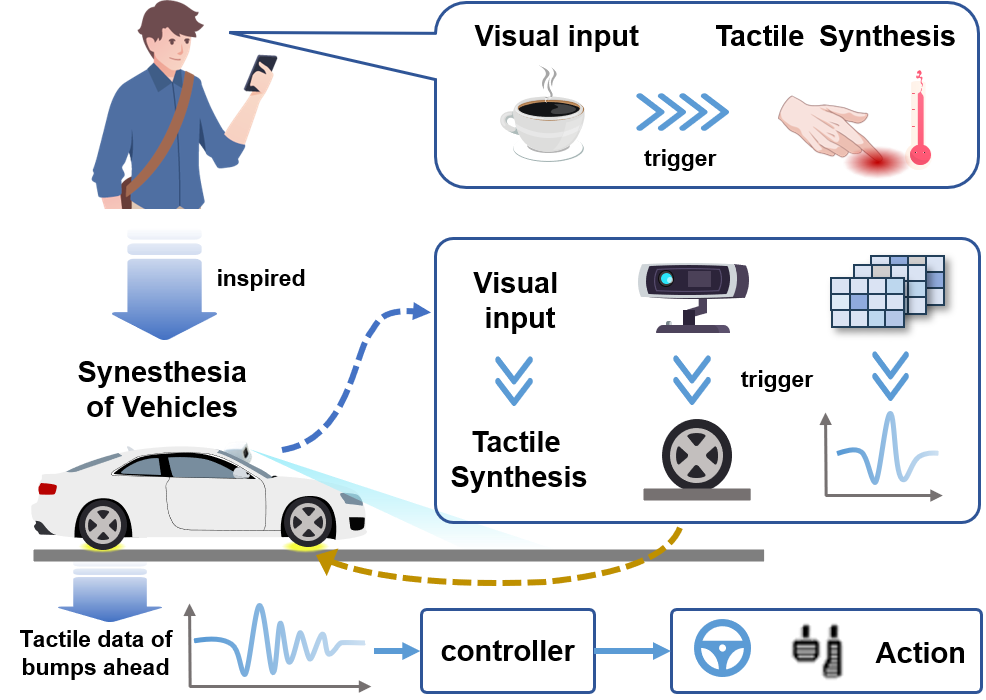}
    \caption{Synesthesia of Vehicles: inspired by human's Synesthesia, providing tactile perception of road ahead and helping vehicles to react proactively.}
    \label{fig:innovation}
\end{figure}

Beyond multi-modal fusion, human synesthesia—where one sensory input triggers another, such as perceiving colors from sounds \cite{synesthesia, soundcolor}—inspires a novel approach. This paper introduces the Synesthesia of Vehicles (SoV), shown in Fig. \ref{fig:innovation}, predicting tactile excitations from forward-looking road images without manual annotations. By establishing a synesthetic mapping between visual and tactile modalities, SoV eliminates spatiotemporal gaps, allowing vehicles to anticipate road excitations and adjust vehicle control timely. For AVs without tactile sensors, this framework provides a realistic tactile dimension. Moreover, real-time tactile data can validate generated excitations, enabling online optimization and self-learning without manual calibration.

To achieve this, we developed a visual-tactile perception system on a real vehicle, collecting multi-modal data across diverse roads and lighting conditions. A spatiotemporal alignment method unifies visual and tactile representations. In addition, a visual-tactile synesthetic (VTSyn) generation model synthesizes high-quality tactile data from visual inputs. Extensive experiments confirm VTSyn’s superior performance, enhancing AV safety through proactive tactile perception.

\section{RELATED WORKS}
Inspired by human synesthesia, this paper proposes a synesthetic perception framework for autonomous vehicles (AVs). This section briefly introduces the definition and principles of human synesthesia, its applications in robotics and industry, and an overview of cross-modal generation methods.
\subsection{Synesthesia}
\subsubsection{Synesthesia of Humans}
Synesthesia is a phenomenon where one sensory stimulus triggers an additional sensory experience, such as sound evoking colors (chromesthesia) \cite{synesthesia, soundtocolor}, sound inducing tactile sensations (auditory-tactile synesthesia) \cite{audiototactile}, or letters/numbers associated with colors (grapheme-color synesthesia) \cite{graphtocolor}. The underlying mechanisms remain partly unclear. One theory suggests cross-connections between brain regions, e.g., between text recognition and the color-processing area V4, may cause synesthesia \cite{causesynaesthetic1, causesynaesthetic2}. Another hypothesis proposes it results from disinhibited feedback or reduced inhibition in neural pathways \cite{causesynaesthetic3}. Cytowic and Eagleman support this, noting acquired synesthesia in non-synesthetes due to conditions like temporal lobe epilepsy or stroke \cite{causesynaesthetic4}.
\subsubsection{Synesthesia of Robots}
Advancements in embodied intelligence have inspired robotic synesthetic perception. Bai et al. introduced the Synesthesia of Machines (SoM) concept \cite{SoM}, while Xiang Cheng et al. developed the SynthSoM dataset for multi-modal sensing \cite{SoM2}. Yang et al. proposed an SoM-enhanced precoding paradigm for vehicular networks \cite{SoM3}. Yuan et al. explored robot synesthesia with point-cloud-based tactile representations for improved action reasoning \cite{robotsyn}, and Zhang et al. developed the Olfactory-Taste Synesthesia Model (OTSM-MBR) for flavor recognition \cite{Olfactory-Taste}. Inspired by human's synesthesia, synesthesia of vehicles can advance AVs through multi-modal sensor fusion, integrating visual and tactile data to enhance environmental understanding and response.
\subsection{Cross-Modal Generation}
The cross-sensory triggering mechanism of synesthesia is similar to cross-modal generation tasks in the field of deep learning, which produces high-quality data (e.g., images, tactile signals) from another modality \cite{crossmodal}. Main methods of cross-modal generation include Generative Adversarial Networks (GANs), Variational Autoencoders (VAEs), and diffusion models.

\textbf{Generative Adversarial Networks (GANs):} introduced by \cite{gan}, GANs feature a generator creating samples and a discriminator evaluating them, evolving into variants like Conditional GAN (CGAN) \cite{cgan} and Pix2Pix \cite{pix2pix}. In robotics, \cite{ganNeRF} used CGAN with NeRF for tactile data generation, and \cite{cganvisual2tactile} employed CGAN with perceptual losses for visual-tactile synthesis.However, despite their high-quality outputs, GANs suffer from limited sample diversity, training instability, mode collapse, and slow convergence\cite{crossmodal}.

\textbf{Variational Autoencoders (VAEs):} encoder-decoder models, earn data distributions by mapping inputs to a probabilistic latent space and reconstructing them to approximate the true distribution\cite{Bidirectional}. Zhu et al. proposed MS-VAE model to enhance cross-modal learning with multiple encoders \cite{msvae}. Although VAEs offer rich latent representations, their outputs are lower quality than other methods. They serve as conditional encoders, as in \cite{cmavae}, combining with GANs for acceleration data generation from visuals.

\textbf{Diffusion Models:} use a forward process to turns data into Gaussian noise and a reverse process (e.g., U-Net) to denoise and reconstruct the original data\cite{DDPM}. Building on Denoising Diffusion Probabilistic Models (DDPMs), Latent Diffusion \cite{latentdm} employs an encoder to map data to a latent space for conditional generation, while \cite{CDCD} and \cite{D2C} enhance audio and image synthesis using diffusion models. In robotics, Yang et al. used latent diffusion for visual-tactile conversion \cite{dftactile}, and Dou et al. generated tactile signals from RGB-D images \cite{tactileRF}. For industrial applications, \cite{DF-CDM} enhanced diffusion with conditional inputs. Their strength in capturing input-condition relationships inspires our use of diffusion for visual-tactile synesthesia in AVs. Inspired by excellent generation capability diffusion models, our study employs a diffusion architecture to achieve visual-tactile synesthetic perception in autonomous vehicles.

\section{Methodology}
\label{Sec:Proposed}

\subsection{Architecture}
\begin{figure*}[htp]
    \centering
    \includegraphics[width=13cm]{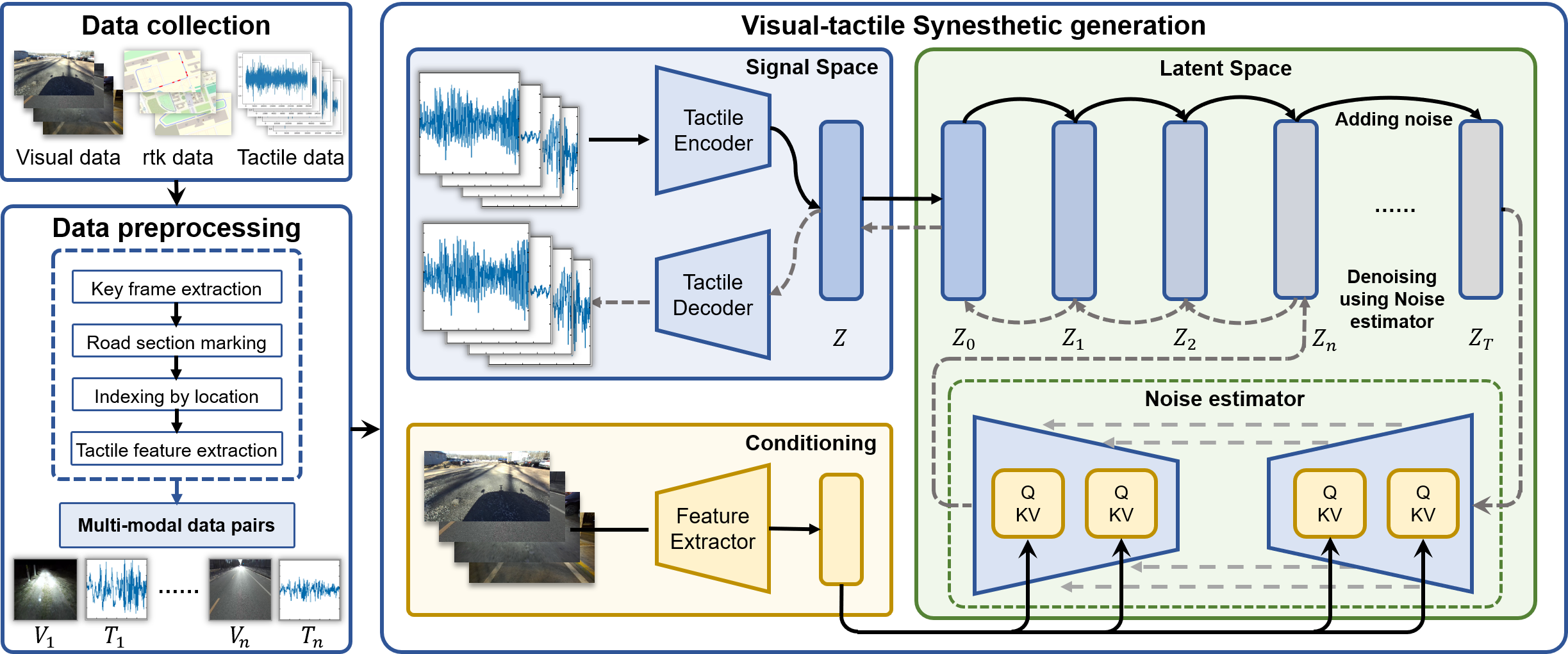}
    \caption{Visual-tactile synesthesia model for cross-modal generation.}
    \label{fig:model}
\end{figure*}

The framework of proposed Synesthesia of Vehicles (SoV), are shown in Fig. \ref{fig:model}. This framework includes data collection, preprocessing with spatiotemporal alignment, and VTSyn generation. Based on the proposed SoV framework, vehicles can sense road surface tactile stimuli in advance during driving, optimizing dynamic control to enhance safety and comfort. Details are introduced as follows.

\subsection{Data collection}
\begin{figure}[htp]
    \centering
    \includegraphics[width=8cm]{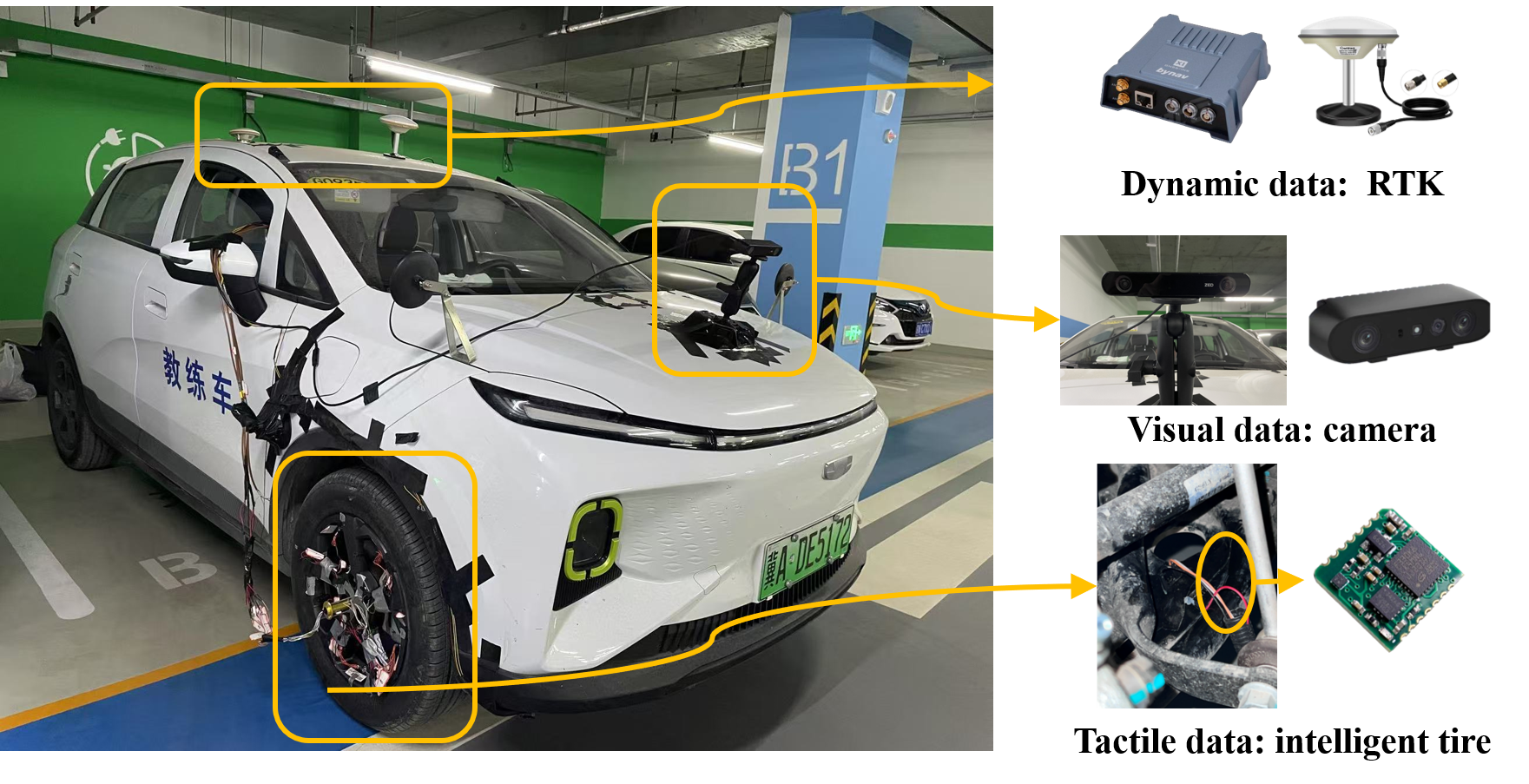}
    \caption{ The proposed visual-tactile perception system for autonomous vehicles. }
    \label{fig:vehicle}
\end{figure}

In this section, we developed a visual-tactile perception system for AVs to collect multi-modal data, as shown in Fig. \ref{fig:vehicle}. The system is implemented on a Geely Geometry E vehicle. A high-resolution ZED 2 camera (30 fps) is mounted on the hood to capture road features, while an intelligent tire with a three-axis ADXL375 accelerometer (500 Hz) on the front right wheel serves as the tactile perception module, recording road-induced excitations. Additionally, a Beiyun M20D RTK module (20 Hz) with antennas collects speed and position information of the vehicle. The Miivii TECH Apex AD10 device is responsible for executing multimodal data acquisition functions.

\begin{figure}[htp]
    \centering
    \includegraphics[width=8cm]{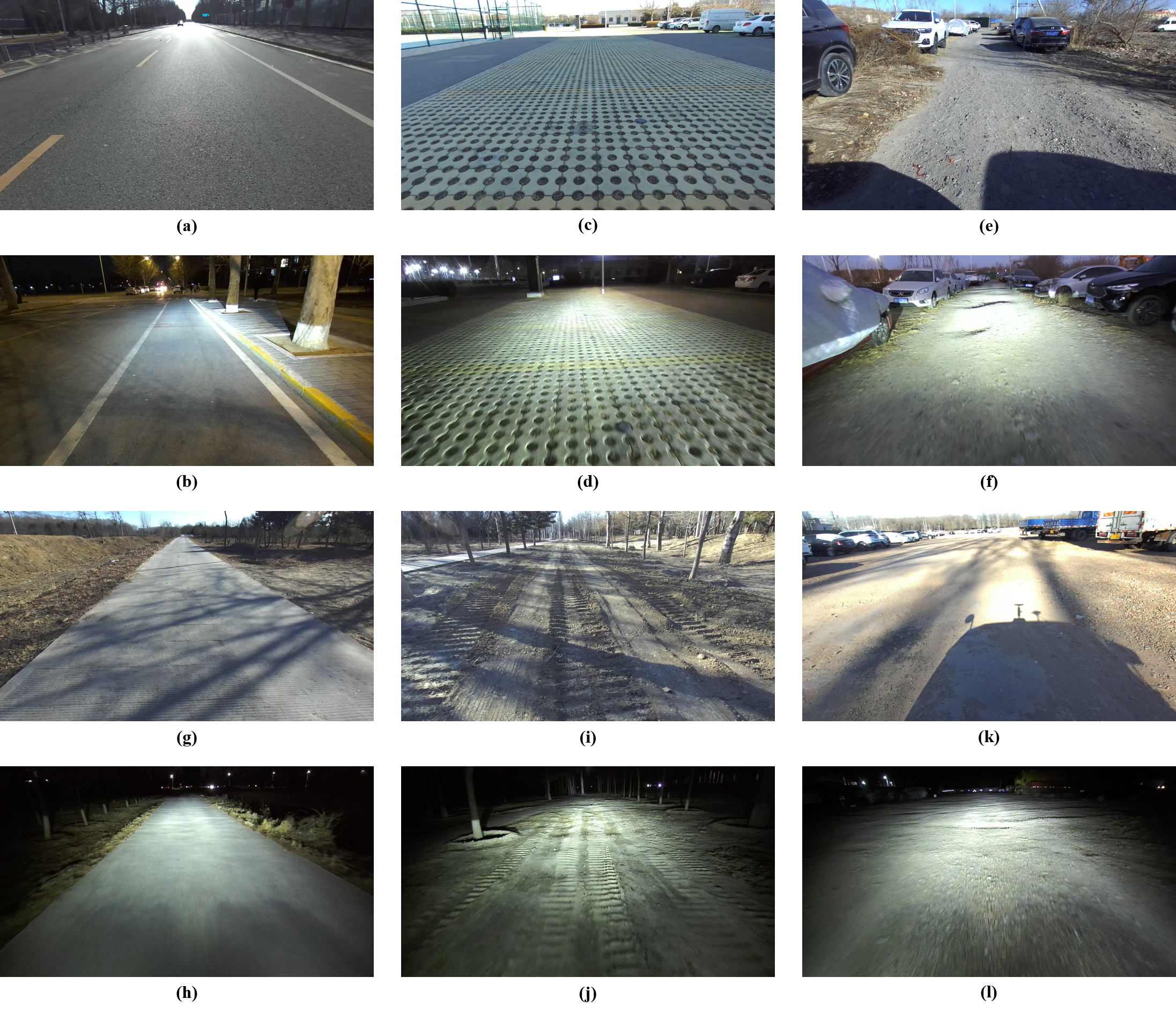}
    \caption{The visual data of different roads on daytime and nighttime: (a)-(b): asphalt road at daytime and nighttime; (c)-(d): brick road at daytime and nighttime; (e)-(f): dirt road at daytime and nighttime; (g)-(h): cement road at daytime and nighttime; (i)-(j): muddy road at daytime and nighttime; (k)-(l): gravel road at daytime and nighttime. }
    \label{fig:visual}
\end{figure}

\begin{figure}[htp]
    \centering
    \includegraphics[width=8cm]{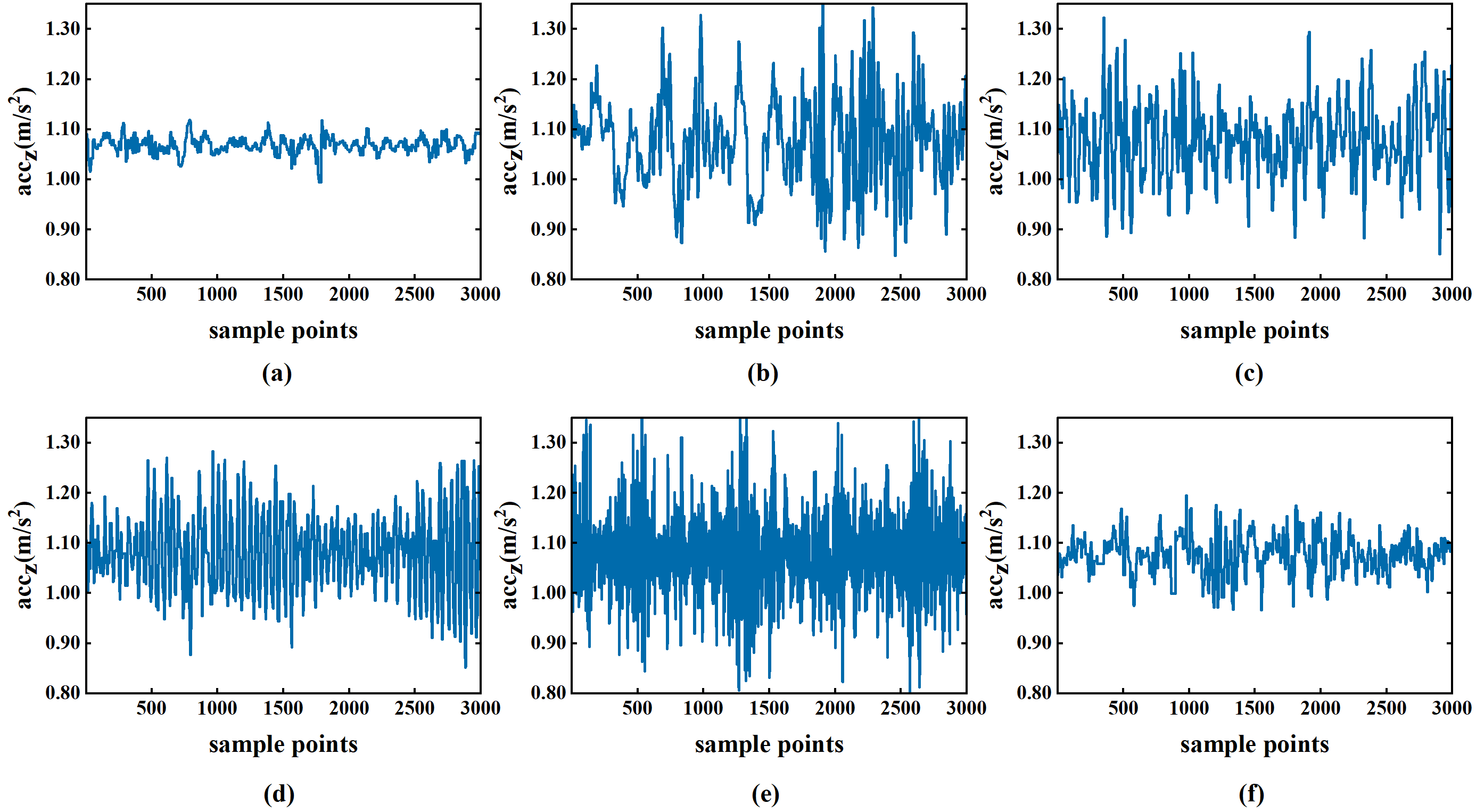}
    \caption{The tactile data corresponding to different roads: (a) asphalt roads; (b) muddy roads; (c) gravel roads; (d) brick roads; (e) dirt roads; (f) cement roads. }
    \label{fig:tactile}
\end{figure}

To build an autonomous driving multi-modal dataset, we collected data from various road surfaces under different lighting and operating conditions. Fig. \ref{fig:visual} illustrates the visual data of six road types—asphalt, concrete, muddy, gravel, brick, and dirt—under daytime and nighttime. Fig. \ref{fig:tactile} presents tactile data from the intelligent tire. Table \ref{table:dataset} presents the statistical information of the collected visuo-tactile multi-modal dataset, displaying the number of multi-modal data pairs across different road types and lighting conditions.

\begin{table}[]
\centering
\caption{Number of the Visual-Tactile Multimodal Dataset }
\label{table:dataset}
\begin{tabular}{ccc}
\hline
\multicolumn{1}{c}{\textbf{Road Type}} & \multicolumn{1}{c}{\textbf{Light Condition}} & \multicolumn{1}{c}{\textbf{Number of Data}} \\ \hline
\multirow{2}{*}{Muddy Road}            & Day                                      & 807                                                     \\ 
                                      & Night                                    & 910                                                     \\ \hline
\multirow{2}{*}{Gravel Road}                            & Day                                      & 730                                                     \\ 
                                      & Night                                    & 1,039                                                   \\ \hline
\multirow{2}{*}{Asphalt Road}                           & Day                                      & 628                                                     \\ 
                                      & Night                                    & 761                                                     \\ \hline
\multirow{2}{*}{Brick Road}                             & Day                                      & 662                                                     \\ 
                                      & Night                                    & 1,014                                                   \\ \hline
\multirow{2}{*}{Dirt Road}                              & Day                                      & 966                                                     \\ 
                                      & Night                                    & 1,730                                                   \\ \hline
\multirow{2}{*}{Cement Road }                           & Day                                      & 939                                                     \\ 
                                      & Night                                    & 1,930                                                   \\ \hline
\end{tabular}
\end{table}

\subsection{Spatiotemporal alignment method for visual and tactile data}

\begin{figure}[htp]
    \centering
    \includegraphics[width=8cm]{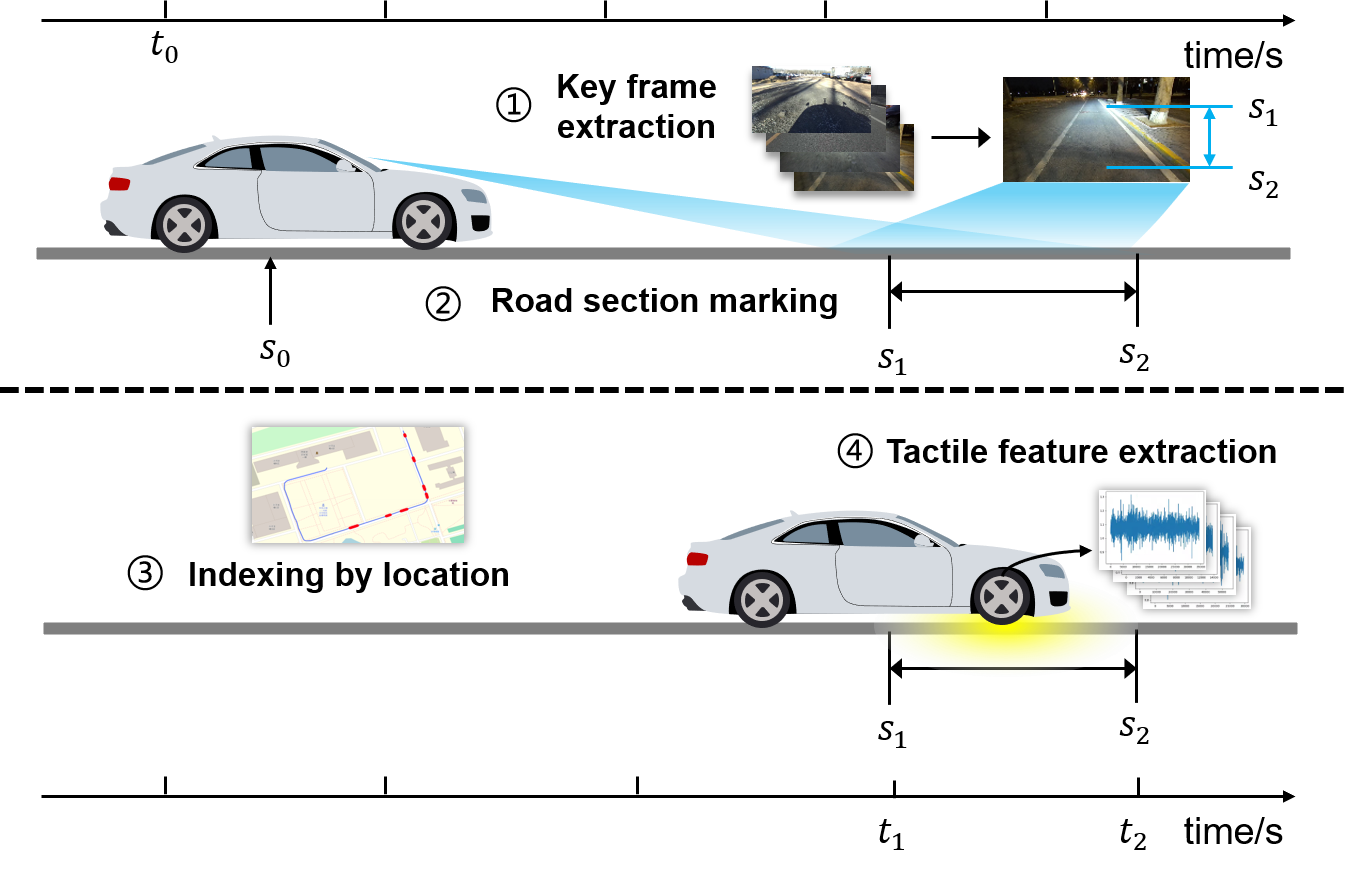}
    \caption{ The procedure of visual-tactile modality alignment algorithm of autonomous vehicles. }
    \label{fig:align}
\end{figure}

This algorithm is structured into four key components: keyframe extraction, target road segment marking, integration of positional information for indexing, and tactile data extraction. The detailed procedure is illustrated in the Fig.~\ref{fig:align}.

\textbf{Keyframe Extraction}: The visual data is acquired at the vehicle's current position, while capturing features of roads ahead, serving as the raw visual data $X_{V}$. We capture visual data at a frequency of 30 fps and extract each frame individually. Next, we will use the visual data as a foundation to extract the corresponding tactile data for each frame of image data.


\textbf{Target Road Segment Marking}: Prior to data collection, the camera was adjusted such that the effective part of the roads in the image is as much as possible. Through actual measurement, the effective road surface segment in the image are located approximately 0.6 to 20 meters ahead of the vehicle. This corresponds to a road segment spanning from $s_1=s_0 + 0.6 $ m to $s_2=s_0 + 20$ m, where $s_0$ represents the current position coordinate of the vehicle. We record the current vehicle position $s_0$ using the RTK device, which allows us to deduce the effective road section present in the visual data.

\textbf{Indexing by location}: The RTK device simultaneously collects the vehicle's forward speed information. By combining the speed and position data, we can estimate the time when the vehicle reaches the road segments shown in the current image. The time required to travel from the current position to the target segment is calculated through speed integration by Eq. \ref{Eq:t1} and \ref{Eq:t2}. The time $t_1$ and $t_2$ is the time when the vehicle travels at $s_1$ and $s_2$ m.

\begin{equation}
\label{Eq:t1}
s_{1}-s_{0}=\int_{t_{0}}^{t_{1}} v d t
\end{equation}

\begin{equation}
\label{Eq:t2}
s_{0}-s_{0}=\int_{t_{0}}^{t_{2}} v d t
\end{equation}


\textbf{Tactile Data Extraction}: Subsequently,  the tactile data corresponding to the vehicle's traversal of the target road segment $s_1$ to $s_2$ m is extracted, represented as $X_{T}^{raw}$, and $fs$ is the sampling rate of tactile data

\begin{equation}
\label{Eq:interpolate}
X_{T}^{raw}=[x_{t1},x_{t1+1/fs},x_{t1+2/fs},\cdots, x_{t2}]
\end{equation}


Additionally, accounting for variations in vehicle speed that may result in inconsistent tactile data lengths, the temporal indexing of tactile data needs to be converted to spatial indexing. We first constructed an interpolation function $f_{interp}$ to fit the extracted raw tactile data to get the resampling tactile data $X_{T}$. Thereby we get the visual-tactile multi-modal data pairs $[X_V, X_T]$ at time $ t_0 $.




\subsection{visual-tactile synesthetic generation model (vtSyn)}


Latent Diffusion is a deep learning algorithm based on VAE and diffusion models, generating high-quality images from random inputs by progressively removing noise\cite{latentdm}. Inspired by this, this paper proposes a vision-tactile synesthesia generation algorithm (vtSyn) for AVs. The details are introduced as follows.

\subsubsection{VAE} In the vtSyn, a variational Autoencoder (VAE) is designed for tactile modality. The encoder comprises four convolutional layers (ConvLayer), each consisting of a 1D convolutional layer (Conv1D), a batch normalization layer (BatchNorm), a ReLU activation function, and a one-dimensional residual connection. After the second ConvLayer, a 1D self-attention mechanism (SelfAttention1D) is incorporated to capture global dependencies within the sequence, improving the model's ability to model long-range features\cite{transformer}.  The formula of SelfAttention1D is shown in Eq. \cref{eq:transpose,eq:output}.

\begin{align}
x^{T} &= \text{transpose}(x_{in}, 1, 2) \label{eq:transpose} \\
x_{ln} &= \text{LayerNorm}(x^{T}) \label{eq:layernorm} \\
x_{\text{attn}} &= \text{mha}(x_{ln}, x_{ln}, x_{ln}) + x^{T} \label{eq:attention} \\
x_{\text{ffn}} &= \text{FFN}(x_{\text{attn}}) + x_{\text{attn}} \label{eq:ffn} \\
x_{out} &= \text{transpose}(x_{\text{ffn}}, 1, 2) \label{eq:output}
\end{align}


At the end of the encoder, a 1D convolutional layer (Conv1D) is used to extract features and adjust the length of the latent features. The decoder's structure mirrors the encoder, with conv1d replaced by transconv1d, and a final tanh activation to output reconstructed data. Further details on the decoder's structure are omitted for brevity.




\subsubsection{Feature extractor} We designed a visual feature extractor to capture deep visual features and input into reverse process as conditions. This study employs a pre-trained ResNet18 model as the feature extractor\cite{resnet}, with its final layer replaced by a fully connected layer to output a feature vector of a specified dimension. In this paper, we set the output dimension of feature extractor is 256.

\subsubsection{Diffusion process}
The diffusion model aims to learn the conditional distribution $ p(\mathbf{x} \mid \mathbf{c}) $, mapping source samples $\mathbf{c}$ (visual features from a feature extractor) to target samples $\mathbf{x}$ (latent tactile features). During training, latent tactile features undergo a forward noise addition process, followed by reverse denoising, with a denoising network $ f_\theta $ trained to predict noise across time steps \cite{DDPM}.

\textbf{Forward process} The forward process is a Markov chain that iteratively adds Gaussian noise to latent features from a Variational Autoencoder (VAE), producing a noisy sample $\mathbf{x}_t$.  Starting from the original data $\mathbf{x}_0$, the noisy sample at step $ t $ is:




\begin{equation}
\label{eq:forward_process}
\begin{split}
q(\mathbf{x}_t \mid \mathbf{x}_0) &= \mathcal{N}(\mathbf{x}_t \mid \sqrt{\bar{\alpha}_t} \mathbf{x}_0, (1 - \bar{\alpha}_t) \mathbf{I}), \\
\mathbf{x}_t &= \sqrt{\bar{\alpha}_t} \mathbf{x}_0 + \sqrt{1 - \bar{\alpha}_t} \mathbf{z}_t, \\
\mathbf{z}_t &\sim \mathcal{N}(\mathbf{0}, \mathbf{I})
\end{split}
\end{equation}

where $\bar{\alpha}_t = \prod_{s=1}^t \alpha_s$. $ T $ is the number of diffusion steps, and $\alpha_1, \ldots, \alpha_T \in (0, 1)$ are hyperparameters controlling noise variance. At $ t = T $, $\mathbf{x}_T$ approximates a standard normal distribution $\mathcal{N}(\mathbf{0}, \mathbf{I})$. We adopt a linear noise schedule with $\beta_{\text{start}} = 10^{-4}$, $\beta_{\text{end}} = 0.02$, and $\text{noise\_steps} = 1000$, where $\alpha_t = 1 - \beta_t$.

\textbf{Reverse process} The reverse process infers the original sample $\mathbf{x}_0$ from the noisy $\mathbf{x}_T \sim \mathcal{N}(\mathbf{0}, \mathbf{I})$. Introducing conditional features $\mathbf{c}$ (visual data), the process is expressed as:


\begin{equation}
\label{eq:Reverse_process}
\begin{aligned}
p_\theta(\mathbf{x}_{t-1} \mid \mathbf{x}_t, \mathbf{c}) 
    &= \mathcal{N}(\mathbf{x}_{t-1} \mid \mu_\theta(\mathbf{c}, \mathbf{x}_t, t), \sigma_t^2 \mathbf{I}) \\
p_\theta(\mathbf{x}_{0:T} \mid \mathbf{c}) 
    &= p(\mathbf{x}_T) \prod_{t=1}^T p_\theta(\mathbf{x}_{t-1} \mid \mathbf{x}_t, \mathbf{c})
\end{aligned}
\end{equation}

The network $ f_\theta $ estimates noise $\mathbf{z}_t$, with variance fixed at $1 - \alpha_t$. The mean $\mu_\theta$ is derived as:
\begin{equation}
\label{eq:mu}
\mu_\theta(\mathbf{c}, \mathbf{x}_t, t) = \frac{1}{\sqrt{\alpha_t}} \left( \mathbf{x}_t - \frac{1 - \alpha_t}{\sqrt{1 - \bar{\alpha}_t}} f_\theta(\mathbf{c}, \mathbf{x}_t, t) \right)
\end{equation}
The conditional posterior is:
\begin{equation}
\label{eq:mu}
\mathbf{x}_{t-1} = \frac{1}{\sqrt{\alpha_t}} \left( \mathbf{x}_t - \frac{1 - \alpha_t}{\sqrt{1 - \bar{\alpha}_t}} f_\theta(\mathbf{c}, \mathbf{x}_t, t) \right) + \sqrt{1 - \alpha_t} \mathbf{z}_t
\end{equation}
Accurate noise prediction by $ f_\theta $ restores $\mathbf{x}_0$. The training objective minimizes the mean squared error of noise prediction:


\begin{equation}
\label{eq:loss}
\begin{aligned}
L = \mathbb{E}{t \sim [1, T]} \Bigg[
\mathbb{E}{\mathbf{c}, \mathbf{x}} \Bigg[
\mathbb{E}_{\mathbf{z}t \sim \mathcal{N}(\mathbf{0}, \mathbf{I})}
\left| f\theta(\mathbf{c}, \mathbf{x}_t, t) - \mathbf{z}_t \right|^2
\Bigg] \Bigg]
\end{aligned}
\end{equation}

\begin{figure}[htp]
    \centering
    \includegraphics[width=9cm]{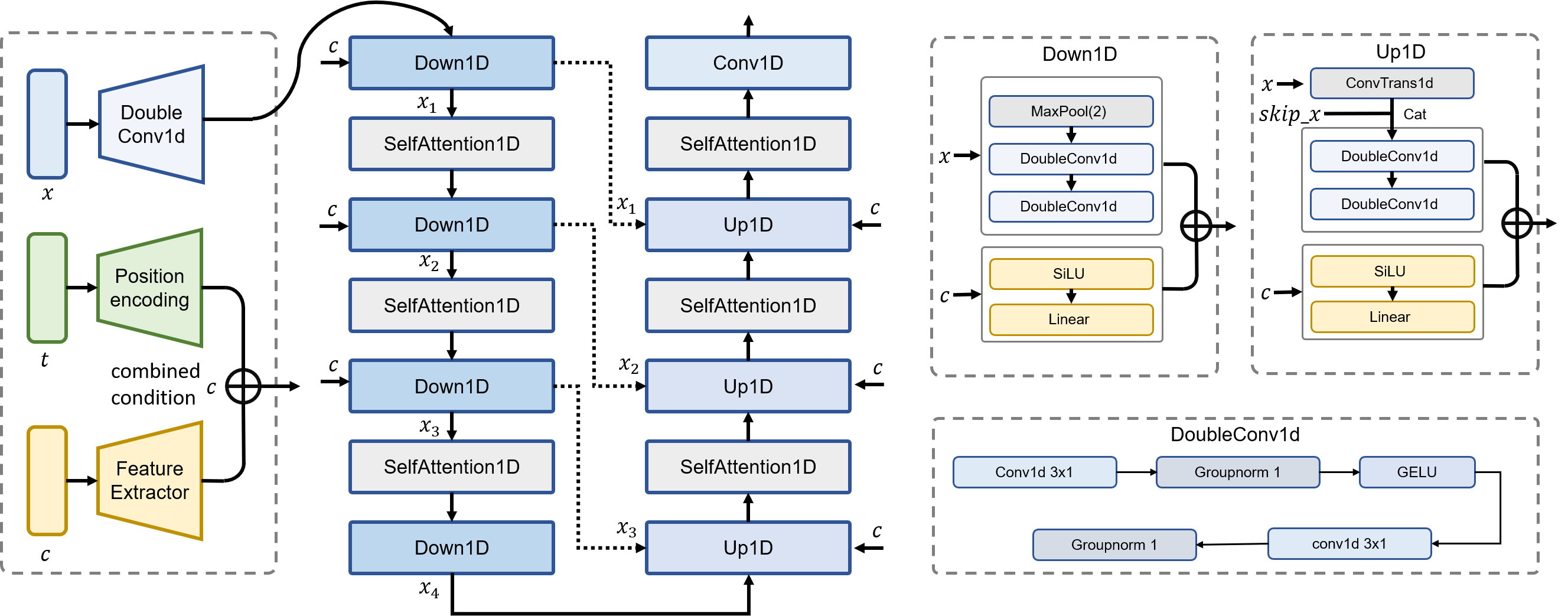}
    \caption{ The denoising network based on u-net architecture. }
    \label{fig:unet}
\end{figure}

U-Net is widely utilized in diffusion models due to its ability to capture multi-scale features during the downsampling process and progressively reconstruct details during the upsampling process, making it effective for predicting noise at each time step\cite{DDPM}. In this paper, we modified the U-net to predict the noise for tactile latent features. The specific structure of the network is as follows as Fig. \ref{fig:unet}.

\section{Experiments and Results}

\subsection{Compared with different baselines}

To quantify the performance of different cross-modal generation models, we computed some evaluation metrics for the generated data, including RMSE and FID. We also defined another frequency-related metrics: freq ssim. The definition are as follows:


\begin{equation}
\label{Eq:freq_sim}
\text{freq }{\text{ssim}} = 1 - \frac{\sum_{i=0}^{K-1} |y_i| \cdot |\hat{y}_i|}{\sqrt{\sum_{i=0}^{K-1} |y_i|^2} \cdot \sqrt{\sum_{i=0}^{K-1} |\hat{y}_i|^2}}
\end{equation}

where: $y_i$  and $\hat{y}_i$ are the FFT results of real and generation tactile signals, $N$ is the signal length, $i$ is the index of the positive frequency components and $ K = \lfloor N/2 \rfloor + 1 $ is the number of positive frequency points.

\begin{table*}[ht]
\centering
\caption{Evaluation metrics for generated data of different algorithms}
\label{Tab:baseline}

{\scriptsize
\begin{tabular}{ccccccccc}
\toprule
\textbf{Metrics} & \textbf{methods} & \textbf{Asphalt} & \textbf{Cement} & \textbf{Muddy Road} & \textbf{Dirt Road} & \textbf{Gravel} & \textbf{Brick Road} & \textbf{All roads} \\
\midrule
\multirow{4}{*}{RMSE} 
  & cgan    & 0.0251 & 0.0406 & 0.1263 & \textbf{0.0920} & 0.1078 & 0.0576 & 0.0755 \\
  & cvae    & \textbf{0.0240} & \textbf{0.0398} & \textbf{0.1216} & 0.0920 & \textbf{0.1070} & \textbf{0.0569} & \textbf{0.0740} \\
  & diffwave & 0.0285 & 0.0496 & 0.1420 & 0.1083 & 0.1286 & 0.0689 & 0.0879 \\
  & ours     & 0.0388 & 0.0629 & 0.1848 & 0.1343 & 0.1667 & 0.0817 & 0.1115 \\
\midrule

\multirow{4}{*}{FID} 
  & cgan    & 113.8707 & 247.9843 & \textbf{70.2881} & 893.6868 & 513.6348 & 139.9504 & 280.7463 \\
  & cvae    & 113.9411 & 305.4583 & 1810.1564 & 1326.4943 & 1604.5078 & 644.5057 & 950.3695 \\
  & diffwave & 106.5137 & 94.8614 & 129.0218 & 313.6005 & 206.4007 & \textbf{60.3242} & 78.1975 \\
  & ours     & \textbf{55.4503} & \textbf{72.4966} & 125.1488 & \textbf{223.5986} & \textbf{182.5980} & 133.4381 & \textbf{53.2676} \\
\midrule

\multirow{4}{*}{freq ssim} 
  & cgan    & 0.6908 & 0.7397 & 0.8024 & 0.7096 & 0.7174 & 0.5862 & 0.7693 \\
  & cvae    & 0.4545 & 0.5273 & 0.3948 & 0.3868 & 0.3572 & 0.5029 & 0.3920 \\
  & diffwave & 0.7776 & 0.7827 & 0.4984 & 0.6194 & 0.8012 & 0.8586 & 0.7725 \\
  & ours     & \textbf{0.8444} & \textbf{0.8399} & \textbf{0.8708} & \textbf{0.8253} & \textbf{0.8189} & \textbf{0.9028} & \textbf{0.7865} \\
\bottomrule
\end{tabular}
}
\end{table*}

Table \ref{Tab:baseline} presents evaluation metrics for different algorithms: CGAN, CVAE, DiffWave\cite{diffwave}, and our proposed vtSyn across six type roads and overall road conditions.

Regarding RMSE, CVAE achieves the lowest values across most road types (e.g., asphalt, cement, muddy, gravel, brick and all roads), indicating its superior point-wise reconstruction accuracy. CGAN performs well on dirt roads, and its overall value (0.0755) exceeds that of vtSyn (0.1115). However, FID results reveal that our vtSyn method exhibits the lowest values on asphalt, cement, dirt, gravel, and overall road conditions, significantly outperforming CGAN’s lowest values (e.g., 70.2881 on muddy roads), suggesting that vtSyn generates data distributions most aligned with true data. DiffWave achieves the lowest FID on brick roads and shows the second best performance on all roads except muddy roads. In terms of freq ssim, vtSyn consistently records the highest values across all road types, highlighting its strong capability in capturing frequency domain structures. DiffWave follows as the second-best performer, while CVAE consistently scores the lowest (e.g., 0.3920 overall). 

In summary, CVAE excels in minimizing local errors but falls short in distribution and frequency similarity, whereas vtSyn demonstrates comprehensive superiority in global distribution and frequency performance, making it highly suitable for generating data across diverse road conditions.

\subsection{visualization of generated tactile data by different methods}

\begin{figure}[htp]
    \centering
\includegraphics[width=9cm]{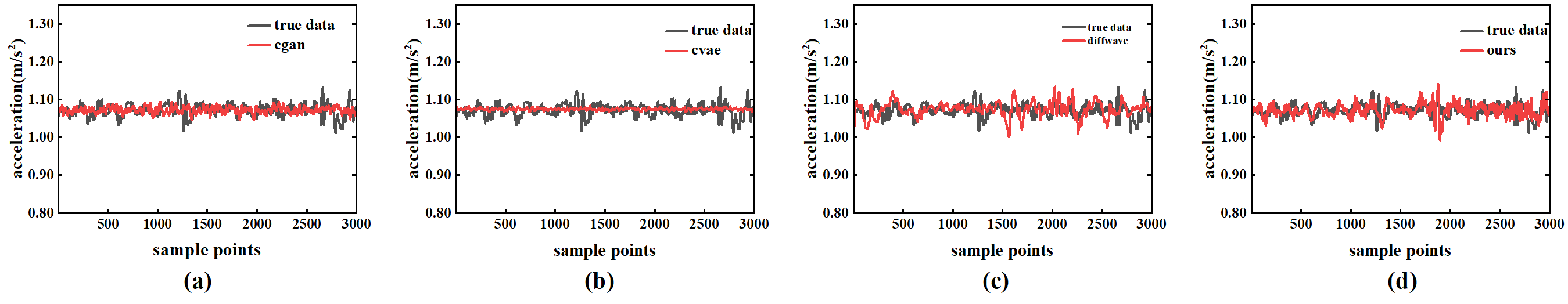}
    \caption{The performance of synthetic data generated by different algorithms on asphalt roads: (a) cgan; (b) cvae; (c) diffwave; (d) ours.  }
    \label{fig:asphalt_icra}
\end{figure}

\begin{figure}[htp]
    \centering
\includegraphics[width=9cm]{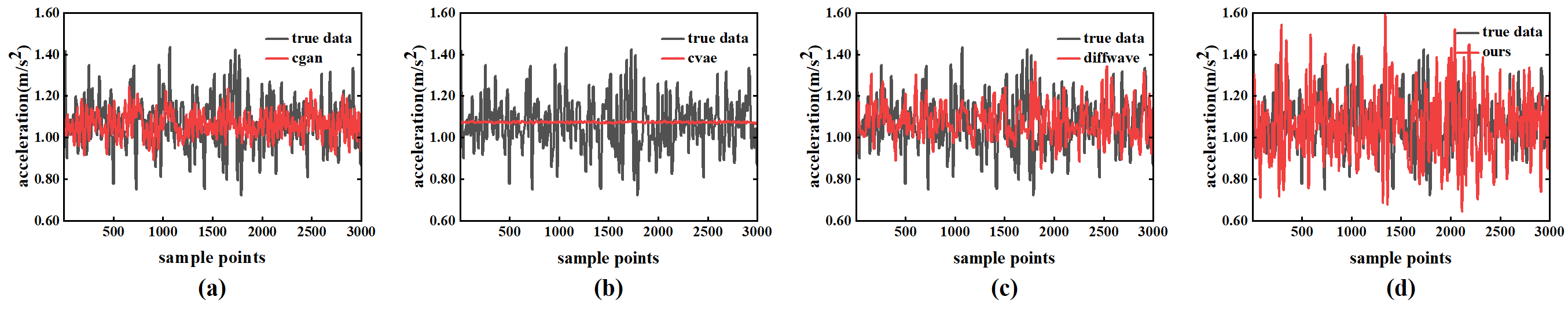}
    \caption{The performance of synthetic data generated by different algorithms on gravel roads:(a) cgan; (b) cvae; (c) diffwave; (d) ours.  }
    \label{fig:gravel_icra}
\end{figure}

To compare the generative performance of different algorithms, we selected the smoothest and roughest road types—asphalt and gravel—from the six investigated, visualizing the generated tactile data, as depicted in Fig. \ref{fig:asphalt_icra} and Fig. \ref{fig:gravel_icra}. The gray line represents the true data, while the red line denotes the pseudo-data generated by each algorithm.

Comparative analysis reveals that, for both asphalt and gravel surfaces, the tactile data generated by our proposed vtSyn model exhibits high similarity to the true data in terms of overall trends and local fluctuations, with consistent curve morphology. As a variant of diffusion models, DiffWave demonstrates good generative performance on both asphalt and gravel surfaces; however, a detailed analysis reveals that the acceleration amplitude on gravel surfaces is slightly lower than the true data. In contrast, the CGAN-generated tactile data for asphalt surfaces exhibits slightly lower amplitudes compared to the true data, while for gravel surfaces, the generated data shows fluctuations within a narrower range than the true data. The CVAE-generated tactile data exhibits substantial deviations from the true data, with a consistently narrow fluctuation range across both road types, failing to reflect the time-frequency characteristics and overall trends of the true data.  

Overall, our proposed vtSyn model achieves the best similarity to true data, validating the superior performance of Stable Diffusion in terms of distribution similarity and error control.

\subsubsection{The range of generated tactile data by different methods}

\begin{figure*}[htp]
    \centering
\includegraphics[width=18cm]{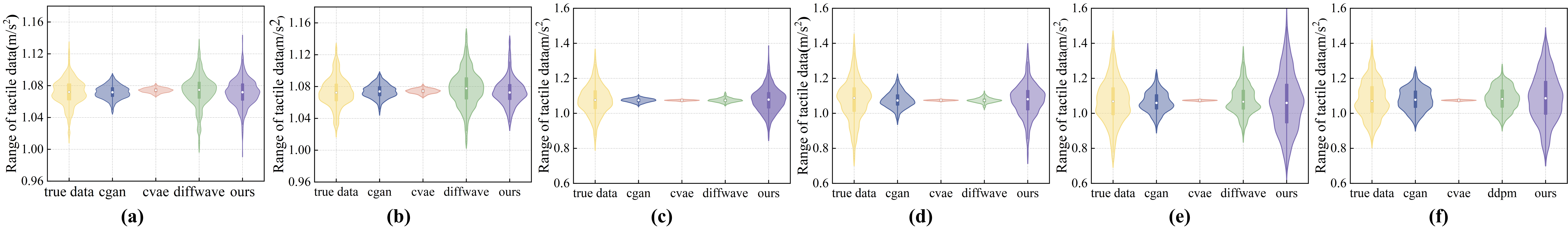}
    \caption{The performance of synthetic data generated by different algorithms on different roads: (a) asphalt road; (b) cement road; (c) dirt road; (d) muddy road; (e) gravel road;(f) brick road;  }
    \label{fig:baseline}
\end{figure*}

The Fig. \ref{fig:baseline} compares the range of tactile data (in m/s²) between synthetic and real data to assess each method's ability of capturing the true tactile distribution.

Comparative analysis reveals that the CGAN method exhibits significant variability in performance across different road surfaces. On asphalt and cement roads, the haptic data generated by CGAN shows a distribution relatively aligned with real data. However, on dirt roads, the generated distribution is overly concentrated, deviating substantially from the true data.  CVAE generates data that is consistently more concentrated than real data across all surfaces, with a narrower amplitude range, indicating its inadequacy in modeling the amplitude fluctuations of haptic data across modalities. DiffWave outperforms CGAN and CVAE on asphalt and cement roads, where the generated data distribution aligns closely with real data. However, on dirt and muddy roads, its synthetic data exhibits excessive concentration and a narrower range, suggesting that while DiffWave's generalization capability across varying road types remains limited. In contrast, our proposed method, vtSyn, demonstrates superior performance across all six road types, effectively approximating the true data distribution. On gravel and brick roads, although some extreme values cause the distribution range to slightly exceed that of the real data, the generated data remains tightly clustered within the true distribution range. 

Overall, vtSyn produces tactile data with a high degree of consistency with real data, maintaining a compact amplitude range and preserving dynamic characteristics.

\subsection{frequency analysis}

\begin{figure*}[htp]
    \centering
\includegraphics[width=18cm]{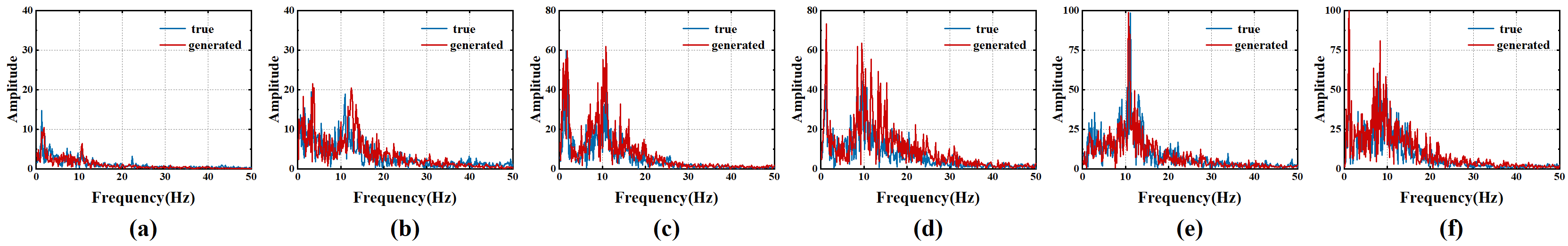}
    \caption{Fast Fourier Transform(fft) analysis of the generation tactile data of vtSyn algorithm at daytime: (a) asphalt roads, (b) cement roads (c) muddy roads, (d) dirt roads, (e)brick roads, (f) gravel roads. }
    \label{fig:fft_day}
\end{figure*}

Fig. \ref{fig:fft_day} shows the frequency domain distributions via  Fast Fourier Transform (FFT) using our vtSyn method and demonstrates consistent alignment between true (blue) and generated (red) signals on six different roads. Notably, low-frequency components (0-20 Hz), which dominate road-induced vibrations, exhibit near-identical distributions, underscoring the model's ability to capture fundamental tactile cues such as surface roughness and vehicle dynamics. Minor discrepancies are observed in higher frequencies ($>30 Hz$), where generated signals occasionally attenuate more rapidly, potentially attributable to perceptual loss functions prioritizing salient features over noise. Overall, the spectral congruence affirms vtSyn's efficacy in synthesizing realistic tactile feedback. Future refinements could incorporate adaptive filtering to mitigate high-frequency artifacts.

In conclusion,  our vtSyn algorithm reveals a high degree of fidelity to real-world vibration patterns across diverse road surfaces. 

\subsection{Downstream task: classification}

To further compare the effectiveness of different cross-modal generation algorithms, we initially trained a CNN-ba classification algorithm based on real tactile data. Subsequently, the tactile data synthesized by various algorithms were evaluated, with the assessment metrics including Accuracy, Precision, Recall, and F1 Score, as presented in the Table.\ref{Table:metrics}. 

The results indicate that our proposed method vtSyn achieved the highest values across nearly all metrics, with an accuracy of approximately 0.64, a recall of around 0.64, and an F1 score of approximately 0.60, significantly outperforming other models. This suggests that our method excels in generating high-quality synthetic samples, likely due to its deterministic flow mechanism, which effectively captures the continuity of the data distribution, thereby enhancing the robustness of downstream classifiers. Benefiting from the substantial advantages of diffusion models in data generation, the DiffWave model achieved the second-best performance, attaining the highest precision value of approximately 0.72, while also ranking second to the proposed vtSyn method in accuracy, recall, and F1 score.

In contrast, the CGAN model exhibited comparatively lower performance, with an accuracy of approximately 0.24, a precision of about 0.15, a recall of around 0.24, and an F1 score of approximately 0.17. This indicates that while CGAN can generate diverse samples through adversarial training, it may introduce noise, leading to a decline in classification accuracy. The tactile data synthesized by the CVAE model performed poorly in classification tasks, with all metrics remaining below 0.10.

Overall, the proposed visual-tactile synesthesia algorithm vtSyn demonstrates the potential to generate high-quality synthetic tactile data and exhibits superior performance in downstream tasks, such as road surface classification.


\begin{table}[ht]
\centering
\caption{Evaluation metrics of different algorithms on road classification tasks}
\label{Table:metrics}
\renewcommand{\arraystretch}{1.5}
\begin{tabular}{ccccc}
\hline
\textbf{Model} & \textbf{accuracy} & \textbf{precision} & \textbf{recall} & \textbf{F1} \\ \hline
cgan           & 0.2481            & 0.1571             & 0.2481         & 0.1715      \\
cvae           & 0.0981            & 0.0224             & 0.0981         & 0.0347      \\
diffwave       & 0.5944            & \textbf{0.6909}     & 0.5944         & 0.5988      \\
ours           & \textbf{0.6406}   & 0.677              & \textbf{0.6406} & \textbf{0.6041} \\ \hline
\end{tabular}
\end{table}

\section{CONCLUSIONS}

This paper proposes the synesthrsia of vehicles framework for autonomous driving, enabling the prediction of road surface tactile excitation from vehicle visual data. A real-vehicle visual-tactile multi-modal perception system and a dataset encompassing diverse road scenarios and lighting conditions are established.  A spatiotemporal alignment method is developed to align visual and tactile representations. Furthermore, a visual-tactile synesthetic generative model (vtSyn), based on a latent diffusion framework, is constructed to achieve unsupervised high-quality tactile data reconstruction. Experimental results demonstrate that the proposed framework delivers high-quality tactile data generation across varying conditions, outperforming existing generative models. Future work will extend to additional scenarios, such as adverse weather conditions like rain and snow, and integrate the framework into autonomous driving systems to optimize downstream tasks, such as vehicle dynamics control. The proposed framework enables proactive tactile perception, significantly enhancing driving safety and offering broad application prospects.



\bibliographystyle{IEEEtran}
\bibliography{reference/ref}
\addtolength{\textheight}{-12cm}

\end{document}